\documentclass[runningheads]{llncs}
\usepackage[T1]{fontenc}
\usepackage{multirow}
\usepackage{graphicx}
\usepackage{cite}
\usepackage{amsmath,amssymb,amsfonts}
\usepackage{algorithmic}
\usepackage{graphicx}
\usepackage{textcomp}
\usepackage[table]{xcolor}
\usepackage{booktabs}
\usepackage{marvosym}

\begin{document}

\title{HGTDP-DTA: Hybrid Graph-Transformer with Dynamic Prompt for Drug-Target Binding Affinity Prediction}

\author{Xi Xiao\inst{1,}\thanks{These authors contributed equally to this work.} \and
Wentao Wang\inst{1,\star} \and
Jiacheng Xie\inst{1} \and
Lijing Zhu\inst{3} \and
Gaofei Chen\inst{1} \and
Zhengji Li\inst{1} \and
Tianyang Wang\inst{1} \and
Min Xu\inst{2,} \thanks{Corresponding author: xu1@cs.cmu.edu}}

\authorrunning{X. Xiao et al.}

\institute{University of Alabama at Birmingham, Birmingham AL 35294, USA \and
Carnegie Mellon University, Pittsburgh PA 15213, USA \and
Bowling Green State University, Bowling Green OH 43403, USA\\
}

\maketitle

\begin{abstract}

Drug target binding affinity (DTA) is a key criterion for drug screening. Existing experimental methods are time-consuming and rely on limited structural and domain information. While learning-based methods can model sequence and structural information, they struggle to integrate contextual data and often lack comprehensive modeling of drug-target interactions. In this study, we propose a novel DTA prediction method, termed HGTDP-DTA, which utilizes dynamic prompts within a hybrid Graph-Transformer framework. Our method generates context-specific prompts for each drug-target pair, enhancing the model’s ability to capture unique interactions. The introduction of prompt tuning further optimizes the prediction process by filtering out irrelevant noise and emphasizing task-relevant information, dynamically adjusting the input features of the molecular graph. The proposed hybrid Graph-Transformer architecture combines structural information from Graph Convolutional Networks (GCNs) with sequence information captured by Transformers, facilitating the interaction between global and local information. Additionally we adopted the multi-view feature fusion method to project molecular graph views and affinity subgraph views into a common feature space, effectively combining structural and contextual information. Experiments on two widely used public datasets, Davis and KIBA, show that HGTDP-DTA outperforms state-of-the-art DTA prediction methods in both prediction performance and generalization ability.

\end{abstract}
\section{Introduction}
In drug development, accurately predicting the binding affinity between a drug molecule and its target protein is a crucial and challenging task~\cite{9305294}. The effectiveness of drug molecules significantly depends on their affinity for target proteins or receptors. However, newly designed drug molecules may sometimes interact with unintended target proteins~\cite{thafar2019comparison}, leading to undesirable side effects. Therefore, it is vital to predict DTA promptly and accurately to ensure the safety and efficacy of new therapeutics.

With the development of high-throughput screening technology, a vast number of molecules that bind to specific targets can be screened more rapidly~\cite{hertzberg2000high}. However, these techniques frequently require expensive chemicals and equipment. The continued advancement of computational technology has allowed drug-target affinities to be predicted more rapidly and affordably through molecular docking methods~\cite{pinzi2019molecular}, which evaluate drug-target interactions using scoring functions~\cite{ballester2010machine}. Despite their efficiency, these docking techniques necessitate various biologically significant pre-processing steps, such as hydrogenation and protonation. In recent years, several machine learning-based approaches have emerged. For instance, KronRLS~\cite{pahikkala2015toward} uses Smith-Waterman similarity representations of the targets and compound similarity-based representations of the drugs. SimBoost~\cite{he2017simboost} constructs new features using drug-drug and drug-target similarities. However, these methods often overlook the structural information embedded in the molecules. To extract more relevant information, deep learning-based methods are being increasingly adopted, particularly string-based and graph-based approaches. Methods like DeepDTA~\cite{ozturk2018deepdta} employ CNNs to learn feature representations from sequence data. Built upon DeepDTA, AttentionDTA~\cite{zhao2019attentiondta} introduces a bilateral multi-headed attention mechanism focusing on key subsequences, which enhances the model's ability to capture sequence features. Nevertheless, these CNN-based models often fail to capture the characteristics of nearby atoms or amino acids. Consequently, graph-based approaches for DTA prediction have been developed, allowing molecules to be represented as graphs~\cite{gong2023ma}. GraphDTA~\cite{nguyen2021graphdta} utilizes GNNs to predict affinities by constructing molecular graphs with atoms as nodes and bonds as edges. DGraphDTA~\cite{jiang2020drug} expands this concept by building molecular topological graphs to represent proteins. MGraphDTA~\cite{yang2022mgraphdta} constructs ultra-deep GNNs with multiple graph convolution layers to capture both local and global structures of compounds. HSGCL-DTA employs hybrid-scale graph contrastive learning to enhance drug-target binding affinity prediction by integrating multi-scale structural information from both drug and target graphs, leveraging contrastive learning to improve feature representations and prediction accuracy.~\cite{ye2023hsgcl} These approaches demonstrate that GNNs can effectively characterize complex molecular structures.

While the above string and graph-based methods have provided relatively high predictive performance, they often struggle with utilizing limited input data effectively. To address these issues, several fusion-based methods have been developed to integrate additional information. FingerDTA~\cite{zhu2022fingerdta} employs molecular fingerprints for drug-target interaction prediction, while MultiscaleDTA~\cite{chen2022multiscaledta} utilizes multiscale graph convolutional networks. HGRL-DTA~\cite{chu2022hierarchical} incorporates hierarchical graph representations, and 3DProtDTA~\cite{voitsitskyi20233dprotdta} leverages 3D protein structures. Additionally, BiComp-DTA~\cite{kalemati2023bicomp} and MSF-DTA~\cite{ma2023predicting} offer comprehensive graph neural network approaches for predicting drug-target affinities. However, these fusion-based methods still exhibit certain limitations, such as inadequate integration of structural and contextual information, reliance on static feature representations, and insufficient noise handling capabilities. To overcome these challenges, we propose a novel approach that incorporates dynamic prompt generation into the DTA prediction task. This mechanism enables the model to generate context-specific prompts, enhancing its ability to capture unique interactions between different drugs and targets. By dynamically adjusting to the specific context of each drug-target pair, our model aims to further improve prediction accuracy and adaptability.

Inspired by the success of prompt in computer vision (CV)~\cite{jia2022visual} and natural language processing (NLP)~\cite{lester2021power, liu2023pre}, we introduce the concept of dynamic prompt generation into the domain of DTA prediction. In CV, prompt-based methods have achieved remarkable success in tasks such as image classification, object detection and segmentation by providing models with context-specific cues that enhance their interpretative capabilities~\cite{jia2022visual}. Similarly, in NLP, prompt techniques have significantly improved performance in tasks like text classification, question answering and machine translation by leveraging prompts to guide the model towards more relevant and accurate responses~\cite{lester2021power}. The similarity between DTA prediction and these fields lies in the need to capture complex, context-specific interactions. Dynamic prompt generation methods can be categorized into two categories: context-specific prompt generation and adaptive prompt tuning. Context-specific prompt generation involves creating prompts tailored to the specific drug-target pair, thus capturing unique interactions more effectively~\cite{rao2022denseclip}. On the other hand, adaptive prompt tuning adjusts the prompts dynamically based on input data characteristics~\cite{zhou2022learning}, thereby increasing the model's flexibility and robustness. By maximizing the relevance of generated prompts to the prediction task, dynamic prompt generation can extract task-relevant information while filtering out irrelevant noise. This approach allows the model to better handle the diverse and complex nature of drug-target interactions. Furthermore, we introduce a new strategy where both the molecular graph view and the affinity subgraph view are projected into a common feature space. This unified feature space facilitates the comprehensive learning of information from both views, ensuring a more accurate and robust prediction~\cite{zhao2017multi}. Therefore, we propose employing dynamic prompt generation and multi-view feature fusion to address the limitations of existing methods and improve the accuracy and robustness of DTA prediction.
The main contributions of this study are summarized as follows:

\textbf{1) Dynamic Prompt Generation}: We introduce dynamic prompt generation into the DTA prediction task for the first time, creating tailored context-specific prompts for each drug-target pair. This mechanism significantly enhances the model’s ability to capture unique interactions, resulting in more discriminative features and improving the overall prediction accuracy while maintaining global structural information.

\textbf{2) Unified Multi-view Feature Fusion}: We propose a novel multi-view feature fusion method that projects the molecular graph view and the affinity subgraph view into a common feature space. This unified feature space effectively combines structural and contextual information, facilitating comprehensive learning and improving the accuracy and robustness of DTA prediction by leveraging the strengths of both views.

\textbf{3) Hybrid Graph-Transformer with Adaptive Feature Enhancement}: We develop a hybrid model that integrates graph convolutional networks (GCNs) to extract structural features from molecular graphs and transformers to capture long-range dependencies in protein sequences. Additionally, we incorporate adaptive feature enhancement, which dynamically adjusts input features to refine the prediction process, ensuring robustness by filtering out irrelevant noise and focusing on critical task-relevant information.

Experimental evaluations conducted on two widely-used public benchmarks across Davis and KIBA demonstrate the effectiveness and superiority of our method compared to the state-of-the-art methods.

\section{Problem Definition}
In the context of drug-target binding affinity (DTA) prediction, the problem is defined as follows~\cite{toropov2005simplified}: Given a set of drugs $D$, a set of protein targets $T$, and a known drug-target affinity matrix $Y \in \mathbb{R}^{|D| \times |T|}$, where each drug $d_i \in D$ is described by a SMILES string and each target $t_j \in T$ is described by its protein sequence, the objective is to predict the unknown binding affinity $\hat{y}_{ij}$ between drug $d_i$ and target $t_j$. This prediction task can be formulated as a regression problem:

\begin{equation}
\hat{y}_{ij} = F_\theta(d_i, t_j)
\end{equation}

where $\hat{y}_{ij} \in \mathbb{R}$ denotes the predicted binding affinity, and $\theta$ represents the learned parameters of the prediction model $F$. The goal is to ensure that the predicted affinity $\hat{y}_{ij}$ closely approximates the true affinity value $y_{ij}$.

\begin{figure*}[htbp]
\centerline{\includegraphics[width=\textwidth]{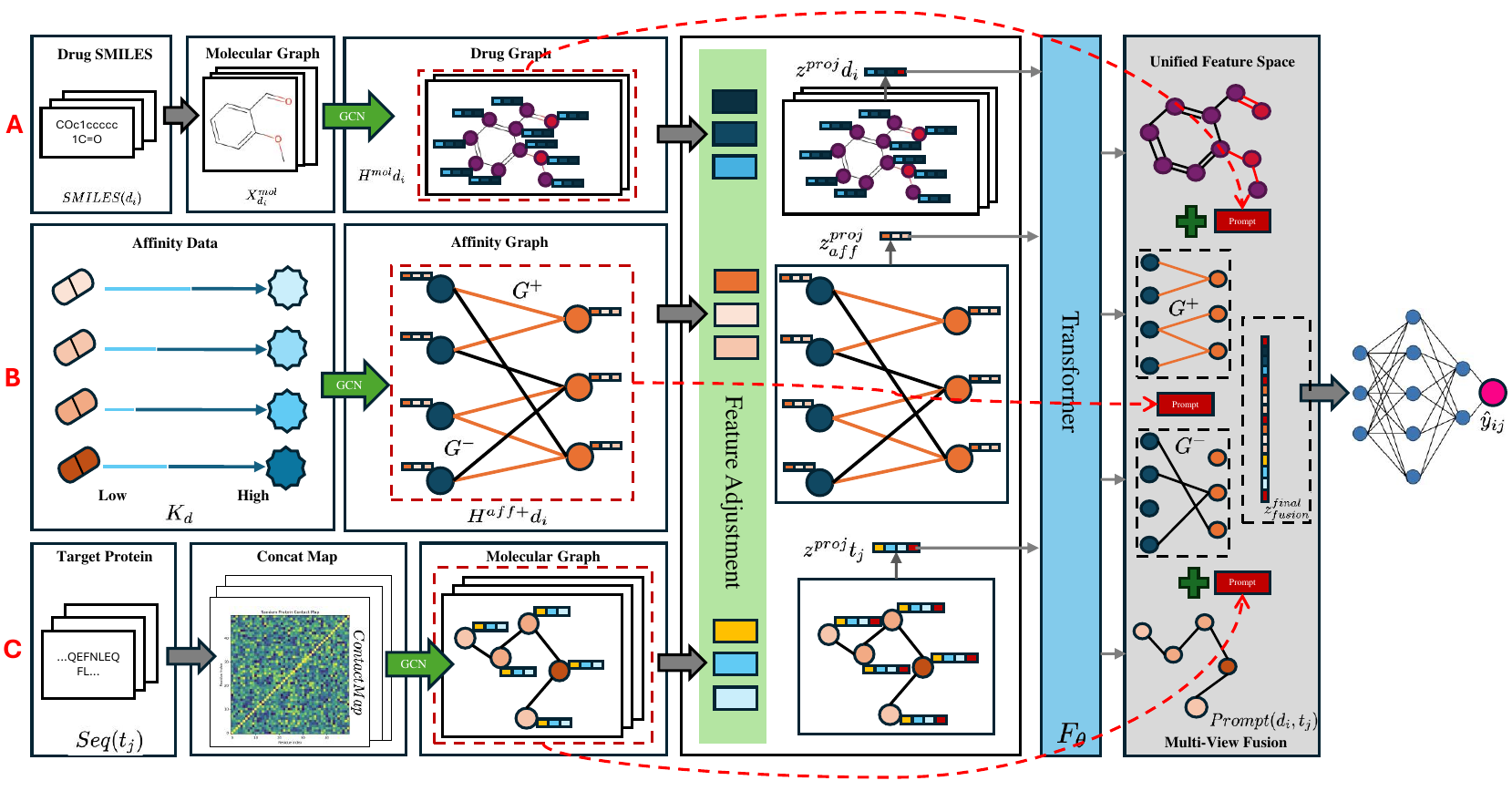}}
    \caption{Overview of the proposed HGTDP-DTA method.}
    \label{fig}
\end{figure*}

\section{Methods}
The overall framework of HGTDP-DTA is illustrated in Fig. 1, which consists of six key modules: dynamic prompt generation, adaptive feature enhancement, drug molecule embedding generation, target protein embedding generation, hybrid Graph-Transformer, and binding affinity prediction. HGTDP-DTA integrates various components to predict drug-target binding affinity effectively. Fig. 1 presents three main paths (A, B, C) in our architecture.


\subsection{Drug Molecule Embedding Generation (Path A)}
Drugs are represented as SMILES strings, which are converted into molecular graphs~\cite{landrum2006rdkit}. Each node in the graph represents an atom, and each edge represents a chemical bond. A three-layer Graph Convolutional Network (GCN) is used to obtain the embeddings of the atom nodes. The initial feature matrix $X^{mol}_{d_i}$ is fed into the GCN to obtain $H^{mol}_{d_i}$, where $h^{mol}_{d_i,v}$ denotes the potential representation of atomic node $v \in V_{d_i}$. The molecular graph embeddings are then combined with the context-specific prompts to form the final drug embeddings (illustrated as $z^{proj}_{d_i}$ in Fig. 2).

\begin{equation}
H^{mol}_{d_i} = \text{GCN}(X^{mol}_{d_i})
\end{equation}

\subsection{Affinity Graph Embedding Generation (Path B)}

Given the known affinity matrix $Y$, a drug-target affinity graph $G = (V, E, W)$ can be constructed, where $V = V_D \cup V_T$ denotes the set of nodes of the drug and target, $E$ denotes the set of edges between drugs and targets, and $W$ is the set of corresponding edge weights. For any $v_i \in V_D, v_j \in V_T, e_{ij} \in E$ denotes the edge between $v_i$ and $v_j$, and $w_{ij} \in W$ denotes the affinity value of drug $v_i$ to target $v_j$.

\begin{equation}
H^{aff+}_{d_i} = \Omega^+(G^+) = \text{ReLU}(\hat{A}^+ \text{ReLU}(\hat{A}^+ X W_0) W_1)
\end{equation}

\begin{equation}
H^{aff-}_{d_i} = \Omega^-(G^-) = \text{ReLU}(\hat{A}^- \text{ReLU}(\hat{A}^- X W_0) W_1)
\end{equation}

where $\hat{A}$ is the regularized adjacency matrix defined as:

\begin{equation}
\hat{A} = \tilde{D}^{-1/2} \tilde{A} \tilde{D}^{-1/2}, \quad \tilde{A} = A + I, \quad \tilde{D} = \text{diag}(\sum_j \tilde{A}_{ij})
\end{equation}

$X$ is the initial feature matrix, $F$ is the dimensionality of the features, $W_0$ and $W_1$ are the weight parameters of the corresponding layers of GCN. $H^{aff+}$ consists of drug node features $H^D_{d_i}$ and target node features $H^T_{t_j}$. After the first layer of GCN encoding, the node can capture information about its one-hop neighbors, i.e., the directly connected heterogeneous nodes. Through the second layer of aggregation, it can obtain information about the same kind of nodes that have the same neighbors as itself.

\begin{equation}
H^{aff+}_{d_i} = \text{GCN}(G^+)
\end{equation}

\begin{equation}
H^{aff-}_{d_i} = \text{GCN}(G^-)
\end{equation}

\subsection{Target Protein Embedding Generation (Path C)}

The target protein is represented as a sequence of amino acids. We convert the protein sequence into a graph where nodes represent residues and edges represent interactions between residues~\cite{jiang2020drug}. Using a similar GCN approach as for drug molecules, we generate the protein embeddings. The initial feature matrix $X^{prot}_{t_j}$ is used to generate $H^{prot}_{t_j}$, and the embeddings are enhanced with context-specific prompts to capture the unique interactions of the target protein (illustrated as $z^{proj}_{t_j}$ in Fig. 2).

\begin{equation}
H^{prot}_{t_j} = \text{GCN}(X^{prot}_{t_j})
\end{equation}

The target protein is represented as a sequence of amino acids, but it can alternatively be represented as a graph with residues as nodes and contacts of residue pairs as edges. To further investigate the hidden intrinsic structural information in the protein sequence, the sequence of protein $t_j$ is converted into a target graph $G^{mol}_{t_j} = (V^{mol}_{t_j}, E^{mol}_{t_j})$, where $V^{mol}_{t_j}$ is the set of protein residues and $E^{mol}_{t_j}$ represents the set of edges between residues. Protein sequence alignment was performed using HHblits~\cite{remmert2012hhblits}, and then PconsC4~\cite{michel2019pconsc4} was used to convert the protein sequence alignment results into a contact map, a residue-residue interaction matrix, whose values represent the Euclidean distance between two residues.

The same encoding is used to generate protein molecule features. After the first layer of GCN, $H^{mol}_{t_j}$ is obtained, and the target node feature $h^{aff+}_{t_j}$ of the positive affinity graph are incorporated into each atom feature:

\begin{equation}
\tilde{h}^{mol}_{t_j,v} = [h^{mol}_{t_j,v} \oplus f(h^{aff+}_{t_j})] \| [h^{mol}_{t_j,v} \ominus f(h^{aff+}_{t_j})]
\end{equation}

After two more layers of GCN and applying the GMP layer to obtain the final representation of all targets:

\begin{equation}
Z^{mol}_{T} = \sum_{t_j \in T} z^{mol}_{t_j}
\end{equation}

\subsection{Hybrid Graph-Transformer Architecture with Multi-view Feature Fusion}
To effectively combine the information from both the molecular graph view and the affinity subgraph view, we propose a unified feature space where both views are projected. This hybrid architecture integrates GCNs and Transformers to capture both local structural information and long-range dependencies. This is achieved through the following steps:

\begin{figure*}[h!]
\centerline{\includegraphics[width=\textwidth]{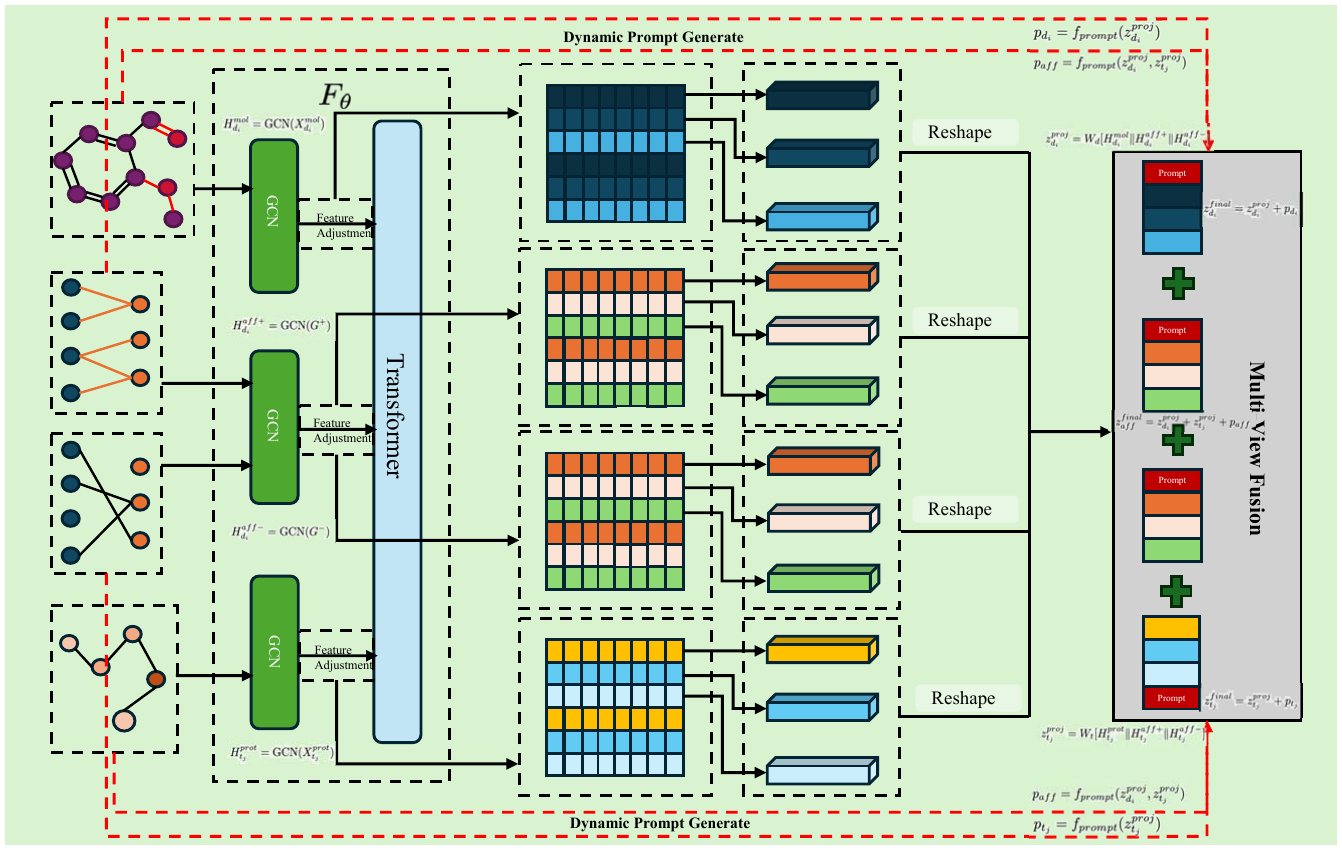}}
\caption{Multi-view Feature Fusion with Prompt Integration in HGTDP-DTA. This module combines features from the molecular graph, affinity graph, and protein graph into a unified feature space, integrating context-specific prompts to enhance prediction accuracy.}
\label{fig:multi_view_feature_fusion}
\end{figure*}

\textbf{Feature Projection:} For each drug $d_i$ and target $t_j$, we obtain their respective embeddings from the molecular graph view ($H^{mol}_{d_i}$, $H^{mol}_{t_j}$) and the affinity subgraph view ($H^{aff+}_{d_i}$, $H^{aff+}_{t_j}$, $H^{aff-}_{d_i}$, $H^{aff-}_{t_j}$). These embeddings are projected into a common feature space using learned projection matrices $W_d$ and $W_t$:

\begin{equation}
z^{proj}_{d_i} = W_d [H^{mol}_{d_i} \| H^{aff+}_{d_i} \| H^{aff-}_{d_i}]
\end{equation}

\begin{equation}
z^{proj}_{t_j} = W_t [H^{prot}_{t_j} \| H^{aff+}_{t_j} \| H^{aff-}_{t_j}]
\end{equation}

\textbf{Dynamic Prompt Generation:} The next step is to generate dynamic prompts based on the projected features. These prompts are designed to capture the specific context of the drug-target interaction, including drug, target, and affinity subgraphs. The process involves a trained prompt generator, which creates context-specific prompts to enhance the feature representations:

\begin{equation}
p_{d_i} = f_{prompt}(z^{proj}_{d_i})
\end{equation}

\begin{equation}
p_{t_j} = f_{prompt}(z^{proj}_{t_j})
\end{equation}

\begin{equation}
p_{aff} = f_{prompt}(z^{proj}_{d_i}, z^{proj}_{t_j})
\end{equation}

\textbf{Prompt Integration:} The final integrated features for the drug, target, and affinity are obtained by combining the projected features with the generated prompts. This integration helps to better capture the unique interactions between the drug and target:

\begin{equation}
z^{final}_{d_i} = z^{proj}_{d_i} + p_{d_i}
\end{equation}

\begin{equation}
z^{final}_{t_j} = z^{proj}_{t_j} + p_{t_j}
\end{equation}

\begin{equation}
z^{final}_{aff} = z^{proj}_{d_i} + z^{proj}_{t_j} + p_{aff}
\end{equation}

The detailed process is illustrated in Fig. \ref{fig:multi_view_feature_fusion}.

\subsection{Drug-target Binding Affinity Prediction}
The final integrated embeddings for the drug, target, and affinity are concatenated and fed into a multi-layer perceptron (MLP) to predict the binding affinity score $\hat{y}_{ij}$:

\begin{equation}
\hat{y}{ij} = MLP(z^{final}{fusion})
\end{equation}

The objective is to minimize the mean squared error (MSE) loss between the predicted affinity and the true affinity:

\begin{equation}
L_{MSE} = \frac{1}{n} \sum_{i=1}^{n} (\hat{y}{ij} - y{ij})^2
\end{equation}

The total loss function combines the MSE loss with the prompt integration loss:

\begin{equation}
L = L_{MSE} + \alpha L_{prompt}
\end{equation}

where $\alpha$ is a hyperparameter to balance the contributions of the different loss components.

\section{Experiments}

\subsection{Datasets}

\textbf{Davis:} The Davis~\cite{davis2011comprehensive} dataset comprises 68 distinct drugs and 442 unique targets, resulting in a total of 30,056 drug-target interactions quantified by $K_d$ (kinase dissociation constant) values. Following previous studies~\cite{davis2011comprehensive}~\cite{pahikkala2015toward}, the $K_d$ values were transformed into logarithmic space as follows:

\begin{equation}
pK_d = -\log_{10}(K_d / 10^9)
\end{equation}

The transformed affinities range from 5.0 to 10.8. For this study, affinities equal to or below 5.0 were considered as negative interactions, indicating very weak or undetectable binding. The drug molecules in this dataset are represented by SMILES strings sourced from the PubChem compound database, and the protein sequences for targets are derived from the UniProt database using gene names and RefSeq accession numbers. The dataset was divided into a training set of 25,746 interactions and a test set of 5,010 interactions.

\textbf{KIBA:} The KIBA~\cite{tang2014making} dataset integrates kinase inhibitor bioactivities from various sources, including $K_i$, $K_d$, and IC50 values, to form a comprehensive measure known as the KIBA score. Originally, the dataset contains 52,498 drugs and 467 targets with a total of 246,088 interaction records. For this study, the dataset was filtered to include 2,111 drugs and 229 targets, resulting in 118,254 drug-target pairs with at least 10 interactions each~\cite{he2017simboost}. The affinities in this dataset range from 0.0 to 17.2, with NaN values indicating missing experimental data. SMILES strings for the drugs were obtained from PubChem, and protein sequences were retrieved from UniProt using corresponding identifiers. The dataset was divided into a training set of 98,545 interactions and a test set of 19,709 interactions.

\subsection{Implementation Details}

All experiments are implemented based on PyTorch. Our models are trained on
a single Nvidia A100 GPU. The same model configurations are utilized on two datasets. We utilized the Adam optimizer to train the entire framework for 2,000 epochs. The batch size is set to 512. The learning rate was set to 5e-4. The dimension of the embedding was set to 128. For the Davis dataset, the threshold $p$ was set to 6, and for the KIBA dataset, it was set to 11. The hyperparameters $\alpha$ and $\beta$ were both set to 0.2, while $\tau$ was set to 0.5.

\subsection{Comparison with State-of-the-art Methods}

We adopt a task-specific paradigm in terms of evaluation metrics in each experiment. Specifically, these metrics include the Mean Squared Error (MSE), Concordance Index (CI), $r^2_m$, and Pearson's Correlation (Pearson). To ensure an unprejudiced comparison and demonstrate the effectiveness of our method, we contrast HGTDP-DTA against
machine learning, sequence, and graph-based methods, along with the fusion-based methods. To make fair comparisons, we obtained the results with convincing parameters for these methods. We report average results using five different random seeds in Table~\ref{tab3} and Table~\ref{tab4}.

\begin{table*}[h!]
\caption{Comparison results to previous state-of-the-art methods on the Davis dataset. Our results are highlighted in \textbf{BOLD}. The best results are highlighted in {\color{blue}BLUE}, while the second-best results are marked in {\color{red}RED}.}
\centering
\begin{tabular*}{\textwidth}{@{\extracolsep{\fill}}llcccccc}
\toprule
\textbf{Type} & \textbf{Method} & \textbf{MSE $\downarrow$} & \textbf{CI $\uparrow$} & \textbf{$r^2_m \uparrow$} & \textbf{Pearson $\uparrow$} \\
\midrule
\multirow{1}{*}{ML-based}
& KronRLS~\cite{davis2011comprehensive} \raisebox{0.5pt}{\color{gray}\scriptsize [NAT BIOTECHNOL'11]} & 0.379 & 0.871 & 0.407 & -  \\
& SimBoost~\cite{he2017simboost} \raisebox{0.5pt}{\color{gray}\scriptsize [CHEMINFORMATICS'17]}  & 0.282 & 0.872 & 0.644 & -  \\
\midrule
\multirow{1}{*}{SQ-based}
& DeepDTA~\cite{ozturk2018deepdta} \raisebox{0.5pt}{\color{gray}\scriptsize [BIOINFORMATICS'18]} & 0.261 & 0.878 & 0.630 & -  \\
& GANsDTA~\cite{zhao2020gansdta} \raisebox{0.5pt}{\color{gray}\scriptsize [FRONT GENET'20]} & 0.276 & 0.881 & - & -  \\
& MRBDTA~\cite{zhang2022predicting} \raisebox{0.5pt}{\color{gray}\scriptsize [BRIEF BIONINFORM'22]} & 0.216 & 0.901 & 0.716 & -  \\
& MFR-DTA~\cite{hua2023mfr} \raisebox{0.5pt}{\color{gray}\scriptsize [BIOINFORMATICS'23]} & 0.221 & 0.905 & 0.705 & -  \\
\midrule
\multirow{1}{*}{GR-based}
& GraphDTA~\cite{nguyen2021graphdta} \raisebox{0.5pt}{\color{gray}\scriptsize [BIOINFORMATICS'21]} & 0.229 & 0.893 & - & -  \\
& DGraphDTA~\cite{jiang2020drug} \raisebox{0.5pt}{\color{gray}\scriptsize [RSC ADV'20]} & 0.203 & 0.904 & \color{blue}0.867 & -  \\
& DeepFusionDTA~\cite{pu2021deepfusiondta} \raisebox{0.5pt}{\color{gray}\scriptsize [ACM TCBI'21]} & 0.253 & 0.887 & - & -  \\
& MGraphDTA~\cite{yang2022mgraphdta} \raisebox{0.5pt}{\color{gray}\scriptsize [CHEM SCI'22]} & 0.207 & 0.900 & 0.710 & -  \\
& DoubleSG-DTA~\cite{qian2023doublesg} \raisebox{0.5pt}{\color{gray}\scriptsize [PHARMACEUTICS'23]} & 0.219 & 0.902 & 0.725 & -  \\
& HSGCL-DTA~\cite{ye2023hsgcl} \raisebox{0.5pt}{\color{gray}\scriptsize [ICTAI'23]} & \color{red}0.155 & \color{red}0.911 & 0.767 & \color{red}0.899  \\
\midrule
\multirow{1}{*}{FU-based}
& FingerDTA~\cite{zhu2022fingerdta} \raisebox{0.5pt}{\color{gray}\scriptsize [BDMA'22]} & 0.234 & 0.895 & 0.678 & -  \\
& MultiscaleDTA~\cite{chen2022multiscaledta} \raisebox{0.5pt}{\color{gray}\scriptsize [METHODS'22]} & 0.200 & 0.889 & 0.738 & -  \\
& HGRL-DTA~\cite{chu2022hierarchical} \raisebox{0.5pt}{\color{gray}\scriptsize [INFORM SCIENCES'22]} & 0.166 & \color{red}0.911 & 0.751 & 0.892  \\
& 3DProtDTA~\cite{voitsitskyi20233dprotdta} \raisebox{0.5pt}{\color{gray}\scriptsize [RSC ADV'23]} & 0.184 & \color{blue}0.917 & 0.722 & -  \\
& BiComp-DTA~\cite{kalemati2023bicomp} \raisebox{0.5pt}{\color{gray}\scriptsize [PLOS COMPUT BIOL'23]} & 0.237 & 0.904 & 0.696 & -  \\
& MSF-DTA~\cite{ma2023predicting} \raisebox{0.5pt}{\color{gray}\scriptsize [JBHI'23]} & 0.194 & 0.906 & - & -  \\
\rowcolor[gray]{0.92}
& \textbf{HGTDP-DTA} \raisebox{0.5pt}{\color{gray}\scriptsize [Ours]} & \textbf{\color{blue}{0.142}} & \textbf{0.910} & \textbf{\color{red}{0.809}} & \textbf{\color{blue}{0.905}}  \\
\bottomrule
\end{tabular*}
\label{tab3}
\end{table*}

\textbf{1) Results on Davis Dataset:} The quantitative results on the Davis dataset are shown in Table~\ref{tab3}. HGTDP-DTA outperforms machine learning-based (ML-based) methods by a large margin, which have limitations in feature extraction and modeling capabilities. The sequence-based (SQ-based) approaches utilize the sequence information of drugs or targets, which could be insufficient in fully capturing the spatial and topological relationships within the molecular structure. Although graph-based (GR-based) methods can capture the topological structure of molecules, they may overlook important features present in the sequence information. Compared to sequence and graph-based methods, our fusion-based (FU-based) HGTDP-DTA leverages both sequence and graph information and shows superior learning ability on almost all evaluation metrics, respectively. Moreover, the success of dynamic prompt generation, unified multi-view feature fusion, and adaptive feature enhancement let HGTDP-DTA overperform other fusion-based methods and achieve a new state-of-the-art.

\begin{table*}[h!]
\caption{Comparison results to previous state-of-the-art methods on the KIBA dataset. Our results are highlighted in \textbf{BOLD}. The best results are highlighted in {\color{blue}BLUE}, while the second-best results are marked in {\color{red}RED}.}.
\centering
\begin{tabular*}{\textwidth}{@{\extracolsep{\fill}}llcccccc}
\toprule
\textbf{Type} & \textbf{Method} & \textbf{MSE $\downarrow$} & \textbf{CI $\uparrow$} & \textbf{$r^2_m \uparrow$} & \textbf{Pearson $\uparrow$} \\
\midrule
\multirow{1}{*}{ML-based}
& KronRLS~\cite{davis2011comprehensive} \raisebox{0.5pt}{\color{gray}\scriptsize [NAT BIOTECHNOL'11]}  & 0.411 & 0.782 & 0.342 & -  \\
& SimBoost~\cite{he2017simboost} \raisebox{0.5pt}{\color{gray}\scriptsize [CHEMINFORMATICS'17]}  & 0.222 & 0.836 & 0.629 & -  \\
\midrule
\multirow{1}{*}{SQ-based}
& DeepDTA~\cite{ozturk2018deepdta} \raisebox{0.5pt}{\color{gray}\scriptsize [BIOINFORMATICS'18]} & 0.194 & 0.863 & 0.673 & -  \\
& GANsDTA~\cite{zhao2020gansdta} \raisebox{0.5pt}{\color{gray}\scriptsize [FRONT GENET'20]} & 0.192 & 0.866 & 0.756 & -  \\
& MRBDTA~\cite{zhang2022predicting}  \raisebox{0.5pt}{\color{gray}\scriptsize [BRIEF BIONINFORM'22]} & 0.146 & 0.892 & 0.778 & -  \\
& MFR-DTA~\cite{hua2023mfr} \raisebox{0.5pt}{\color{gray}\scriptsize [BIOINFORMATICS'23]} & 0.136 & 0.898 & 0.789 & -  \\
\midrule
\multirow{1}{*}{GR-based}
& GraphDTA~\cite{nguyen2021graphdta} \raisebox{0.5pt}{\color{gray}\scriptsize [BIOINFORMATICS'21]} & 0.139 & 0.891 & - & -  \\
& DGraphDTA~\cite{jiang2020drug} \raisebox{0.5pt}{\color{gray}\scriptsize [RSC ADV'20]}  & 0.126 & 0.904 & \color{blue}0.903 & -  \\
& DeepFusionDTA~\cite{pu2021deepfusiondta} \raisebox{0.5pt}{\color{gray}\scriptsize [ACM TCBI'21]} & 0.162 & 0.887 & - & -  \\
& MGraphDTA~\cite{yang2022mgraphdta} \raisebox{0.5pt}{\color{gray}\scriptsize [CHEM SCI'22]} & 0.128 & 0.902 & 0.801 & -  \\
& DoubleSG-DTA~\cite{qian2023doublesg} \raisebox{0.5pt}{\color{gray}\scriptsize [PHARMACEUTICS'23]}  & 0.167 & 0.896 & 0.787 & -  \\
& HSGCL-DTA~\cite{ye2023hsgcl} \raisebox{0.5pt}{\color{gray}\scriptsize [ICTAI'23]}  & \color{red}0.124 & \color{red}0.905 & 0.803 & \color{red}0.907  \\
\midrule
\multirow{1}{*}{FU-based}
& FingerDTA~\cite{zhu2022fingerdta} \raisebox{0.5pt}{\color{gray}\scriptsize [BDMA'22]} & 0.150 & 0.885 & 0.750 & -  \\
& MultiscaleDTA~\cite{chen2022multiscaledta} \raisebox{0.5pt}{\color{gray}\scriptsize [METHODS'22]} & 0.135 & 0.893 & 0.793 & -  \\
& HGRL-DTA~\cite{chu2022hierarchical} \raisebox{0.5pt}{\color{gray}\scriptsize [INFORM SCIENCES'22]}  & 0.166 & 0.899 & 0.789 & \color{red}0.907  \\
& 3DProtDTA~\cite{voitsitskyi20233dprotdta} \raisebox{0.5pt}{\color{gray}\scriptsize [RSC ADV'23]}  & 0.183 & 0.893 & 0.784 & - \\
& BiComp-DTA~\cite{kalemati2023bicomp} \raisebox{0.5pt}{\color{gray}\scriptsize [PLOS COMPUT BIOL'23]} & 0.163 & 0.891 & 0.757 & -  \\
& MSF-DTA~\cite{ma2023predicting}\raisebox{0.5pt}{\color{gray}\scriptsize [JBHI'23]}  & \color{red}0.124 & 0.899 & - & -  \\
\rowcolor[gray]{0.92}
& \textbf{HGTDP-DTA} \raisebox{0.5pt}{\color{gray}\scriptsize [Ours]} & \textbf{\color{blue}{0.119}} & \textbf{\color{blue}{0.912}} & \textbf{\color{red}{0.815}} & \textbf{\color{blue}{0.913}}  \\
\bottomrule
\end{tabular*}
\label{tab4}
\end{table*}

\textbf{2) Results on KIBA Dataset:} To further prove generalizing performances, our model is evaluated on the KIBA dataset, the results are shown in Table~\ref{tab4}. Compared to state-of-the-art methods, our HGTDP-DTA achieves the best performances with 0.119 MSE, 0.912 CI, and 0.913 Pearson, and second-best performance with 0.815 $r^2_m$. Thus, it reveals that our fusion-based HGTDP-DTA is more advantageous to process drug-target affinity data compared to machine learning, sequence, and graph-based models, and consistently outperforms other fusion-based state-of-the-art methods.

\setlength{\tabcolsep}{0.12cm} 
\begin{table}[htbp]
\caption{Ablation studies were conducted on each component using the Davis and KIBA datasets. The components evaluated include Dynamic Prompt Generation (DP), Graph Convolutional Networks (GCN), and Transformer (Trans). The best results are highlighted in bold.}
\label{tab:ablation}
\centering
\begin{tabular}{@{}cccccccccccc@{}}
\toprule
\multicolumn{3}{c}{Module Settings} & \multicolumn{2}{c}{MSE $\downarrow$} & \multicolumn{2}{c}{CI $\uparrow$} & \multicolumn{2}{c}{$r^2_m \uparrow$} & \multicolumn{2}{c}{Pearson $\uparrow$} \\ 
\cmidrule{1-3} \cmidrule{4-5} \cmidrule{6-7} \cmidrule{8-9} \cmidrule{10-11}
DP  & GCN & Trans & Davis & KIBA & Davis & KIBA & Davis & KIBA & Davis & KIBA \\ 
\midrule
& $\checkmark$ & & 0.180 & 0.140 & 0.905 & 0.895 & 0.725 & 0.760 & 0.875 & 0.895 \\
$\checkmark$ & $\checkmark$ & & 0.161 & 0.127 & 0.907 & 0.904 & 0.743 & 0.785 & 0.897 & 0.906 \\
& $\checkmark$ & $\checkmark$ & 0.165 & 0.129 & 0.909 & 0.907 & 0.748 & 0.788 & 0.897 & 0.906 \\
$\checkmark$ & $\checkmark$ & $\checkmark$ & \textbf{0.142} & \textbf{0.119} & \textbf{0.910} & \textbf{0.912} & \textbf{0.809} & \textbf{0.815} & \textbf{0.905} & \textbf{0.913} \\
\bottomrule
\end{tabular}
\end{table}

\subsection{Ablation Study}

To further explore the contribution of the components in our HGTDP-DTA model, we performed a series of ablation experiments. We repeated the experiment five times on the Davis and KIBA datasets to average the results. The results in Table~\ref{tab:ablation} show that the performance of each variant produces increased performance compared to the basic model. The dynamic prompt generation module has the largest influence on the model performance, suggesting that context-specific prompts are crucial for capturing unique interactions between drug-target pairs. The hybrid Graph-Transformer architecture also contributes substantially to the model's accuracy, indicating the importance of combining structural and sequential data.

\subsection{Visualization Analysis}

\begin{figure*}[h!]
\centering
\includegraphics[width=6cm]{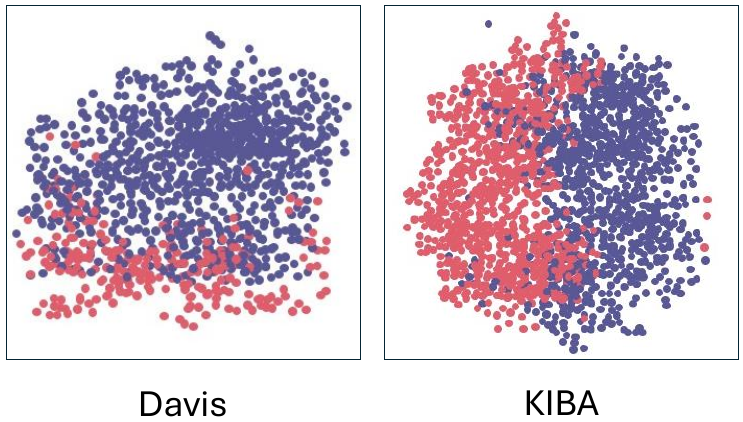}
\caption{Visualization of affinity representations. Red: weak affinity. Blue: strong affinity.}
\label{fig:fitting_performance}
\end{figure*}

In this subsection, we design an additional experiment to explore the representation power of the proposed model from the view of affinity representations.We divide affinities into two clusters using predefined thresholds from previous studies~\cite{davis2011comprehensive, he2017simboost}: a pK\(_d\) value of 7 for the Davis dataset and a KIBA score of 12.1 for the KIBA dataset. Affinities below these thresholds are classified as weak, while those above them are classified as strong. This division is applied to the test sets of both benchmark datasets. We use the trained HGTDP-DTA model to extract learned representations of testing samples before the final prediction layer. This analysis assumes that affinities should cluster closely together for similar interactions and separate distinctly for different ones.Fig. 3 shows the t-SNE visualization of affinity representations for the Davis and KIBA datasets generated by the HGTDP-DTA model. Red points indicate weak affinities, while blue points represent strong affinities. The t-SNE plots reveal a clear mix of strong and weak affinity representations in both datasets, highlighting the model's ability to capture nuanced drug-target interactions.Tables 1 and 2 report the clustering performance of affinity representations for our model and baselines on the two datasets. Our HGTDP-DTA model achieves competitive performance compared to baseline methods. To visualize the affinity representations intuitively, we sample weak and strong affinities in a 1:1 ratio from the KIBA dataset test set and project them into 2D space using t-SNE. As shown in Fig. 4, HGTDP-DTA can effectively distinguish weak affinities (red) from strong ones (blue), indicating that HGTDP-DTA provides more detailed affinity representations, leading to better binding affinity prediction performance.

Overall, the t-SNE visualizations demonstrate that the HGTDP-DTA model effectively learns and distinguishes features of strong and weak affinities. The clear mix and distribution patterns validate the model's ability to accurately predict affinities and capture unique drug-target interactions.

\section{Conclusion}

In this study, we proposed HGTDP-DTA, a novel model for drug-target binding affinity (DTA) prediction. This model integrates dynamic prompts within a hybrid Graph-Transformer framework, combining structural information of molecules with interaction data from drug-target pairs. By capturing features from multiple perspectives, HGTDP-DTA enhances prediction accuracy and robustness. It utilizes Graph Convolutional Networks (GCNs) for processing molecular graphs and Transformers for sequence information, facilitating feature fusion. Dynamic prompts improve the model's ability to capture unique interactions by generating context-specific prompts for each drug-target pair. Our experimental results on benchmark datasets, Davis and KIBA, demonstrate that HGTDP-DTA outperforms state-of-the-art methods in terms of both prediction performance and generalization capability. The ablation studies further emphasize the significant contributions of dynamic prompt generation, GCNs, and Transformer modules to the overall model performance. The hybrid Graph-Transformer architecture effectively combines local structural information and long-range dependencies, providing a comprehensive understanding of drug-target interactions. In conclusion, the integration of known drug-target associations and the use of multi-view feature fusion with dynamic prompts significantly enhance DTA prediction. Future work will explore additional heterogeneous graph embedding techniques, optimize model parameters, and accelerate the training process to further improve the efficiency and effectiveness of the HGTDP-DTA model.

\clearpage

\bibliographystyle{plain} 
\bibliography{reference}
\end{document}